\documentclass[letterpaper, 10 pt, conference]{ieeeconf}  

\IEEEoverridecommandlockouts                              
\usepackage{float}
\usepackage{booktabs} 
\usepackage{threeparttable}
\usepackage{overpic}
\usepackage{graphics} 
\usepackage{subfig} 
\usepackage{wrapfig}

\usepackage{amsthm} 
\usepackage{amsmath,amssymb}

\usepackage{newtxtext}
\usepackage{graphicx}
\usepackage{tabularx}
\usepackage{multirow, makecell}
\usepackage{diagbox}
\usepackage{slashbox}
\usepackage{rotating}
\usepackage{cite}

\usepackage{url}
\usepackage{bm}
\usepackage[table]{xcolor}
\usepackage[hidelinks]{hyperref}
\usepackage{siunitx} 
\usepackage{booktabs}
\usepackage{cleveref}
\usepackage{mathtools} 
\usepackage{lipsum} 

\let\labelindent\relax 
\usepackage{enumitem} 

\usepackage[ruled,vlined]{algorithm2e}

\usepackage{pifont} 

\usepackage[nolist, nohyperlinks]{acronym}
\acrodef{MPC}[MPC]{Model Predictive Control}
\acrodef{QP}[QP]{Quadratic Program}
\acrodef{CBF}[CBF]{Control Barrier Function}

\newcommand{\vu}{{\boldsymbol u}}

\DeclareMathAlphabet{\mathmybb}{U}{bbold}{m}{n}

\DeclareMathOperator*{\argmin}{arg\,min}

\theoremstyle{definition}

\theoremstyle{definition}

\theoremstyle{definition}

\usepackage{fontawesome5}
\usepackage{float}

\makeatletter
\let\save@mathaccent\mathaccent
\newcommand*\if@single[3]{%
  \setbox0\hbox{${\mathaccent"0362{#1}}^H$}%
  \setbox2\hbox{${\mathaccent"0362{\kern0pt#1}}^H$}%
  \ifdim\ht0=\ht2 #3\else #2\fi
  }
\newcommand*\rel@kern[1]{\kern#1\dimexpr\macc@kerna}
\newcommand*\widebar[1]{\@ifnextchar^{{\wide@bar{#1}{0}}}{\wide@bar{#1}{1}}}
\newcommand*\wide@bar[2]{\if@single{#1}{\wide@bar@{#1}{#2}{1}}{\wide@bar@{#1}{#2}{2}}}
\newcommand*\wide@bar@[3]{%
  \begingroup
  \def\mathaccent##1##2{%
    \let\mathaccent\save@mathaccent
    \if#32 \let\macc@nucleus\first@char \fi
    \setbox\z@\hbox{$\macc@style{\macc@nucleus}_{}$}%
    \setbox\tw@\hbox{$\macc@style{\macc@nucleus}{}_{}$}%
    \dimen@\wd\tw@
    \advance\dimen@-\wd\z@
    \divide\dimen@ 3
    \@tempdima\wd\tw@
    \advance\@tempdima-\scriptspace
    \divide\@tempdima 10
    \advance\dimen@-\@tempdima
    \ifdim\dimen@>\z@ \dimen@0pt\fi
    \rel@kern{0.6}\kern-\dimen@
    \if#31
      \overline{\rel@kern{-0.6}\kern\dimen@\macc@nucleus\rel@kern{0.4}\kern\dimen@}%
      \advance\dimen@0.4\dimexpr\macc@kerna
      \let\final@kern#2%
      \ifdim\dimen@<\z@ \let\final@kern1\fi
      \if\final@kern1 \kern-\dimen@\fi
    \else
      \overline{\rel@kern{-0.6}\kern\dimen@#1}%
    \fi
  }%
  \macc@depth\@ne
  \let\math@bgroup\@empty \let\math@egroup\macc@set@skewchar
  \mathsurround\z@ \frozen@everymath{\mathgroup\macc@group\relax}%
  \macc@set@skewchar\relax
  \let\mathaccentV\macc@nested@a
  \if#31
    \macc@nested@a\relax111{#1}%
  \else
    \def\gobble@till@marker##1\endmarker{}%
    \futurelet\first@char\gobble@till@marker#1\endmarker
    \ifcat\noexpand\first@char A\else
      \def\first@char{}%
    \fi
    \macc@nested@a\relax111{\first@char}%
  \fi
  \endgroup
}

\makeatother 
\makeatother

\DeclareCaptionFont{mysize}{\fontsize{8}{9.6}\selectfont}
\title{\LARGE \bf High-Speed Forward and Reverse Tractor-Trailer Driving with Learned Vehicle Dynamics}

\author{Aaron Chen${^*}^{1}$, David Shen${^*}^{1}$, Taekyung Kim${^*}^{2}$, Kaleb Ben Naveed$^{2}$, Dimitra Panagou$^{2,3}$
\thanks{$^*$These authors contributed equally to this work. %
}
\thanks{$^{1}$Department of Electrical Engineering and Computer Science, $^{2}$Department of Robotics, $^{3}$Department of Aerospace Engineering, University of Michigan, Ann Arbor, MI, 48109, USA {\tt\footnotesize \{aach, dyshen, taekyung, kbnaveed, dpanagou\}@umich.edu } }}

\begin{document}
\maketitle
\thispagestyle{empty}
\pagestyle{empty}

\setcounter{dbltopnumber}{2}
\setlength{\dblfloatsep}{8pt plus 1pt minus 1pt}
\setlength{\dbltextfloatsep}{8pt plus 1pt minus 1pt}
\captionsetup[figure]{skip=3pt}
\captionsetup[table]{skip=3pt}

\begin{abstract}
Tractor-trailer systems are difficult to control, especially unstable during reverse motion, and their dynamics depend on difficult-to-measure factors, including road friction and suspension coefficients. We introduce a learned dynamics model for tractor-trailer control and compare it with a reduced-order dynamics model with Fiala tire forces as an analytic baseline.  Specifically, we compare a model predictive path integral (MPPI) controller that uses the learned dynamics model to an MPPI controller that uses the analytic model, as well as to three additional baseline controllers. In the main closed-loop tracking experiments across four simulated environments, MPPI with learned dynamics achieves mean speeds closer to the target magnitudes and lower root-mean-square (RMS) hitch angles than MPPI with the analytic model at every tested setpoint, with higher completion rates in seven of eight reverse configurations.
\href{https://ds12a.github.io/drift-with-trailer/}{\textcolor{red}{[Project Page]}} \footnote{Project page: \href{https://ds12a.github.io/drift-with-trailer/}{https://ds12a.github.io/drift-with-trailer/}} \href{https://github.com/tkkim-robot/drift-with-trailer/}{\textcolor{red}{[Code]}}
\end{abstract}


\section{INTRODUCTION}

Despite advances in autonomous vehicle control~\cite{amini2020learning,lelko2024optimal}, uncertain dynamics remain challenging~\cite{learningmpc}. Particularly, Model Predictive Control (MPC), which uses predictions from a system model to optimize a finite-horizon control sequence subject to input and state constraints, is a common control framework applied to vehicle control. As MPC relies on predicted costs and constraint violations determine the selected inputs, model errors can degrade closed-loop performance~\cite{learningmpc,uncertainty_mpc}.

These challenges are particularly pronounced in tractor-trailer systems, whose coupled dynamics are nonlinear~\cite{KAYACAN20141}. Tractor-trailers are unstable, especially in reverse, and trailer motion must be controlled indirectly through tractor control~\cite{altafini2001feedback}. At high speeds, the controller must anticipate the vehicle, tire, and hitch behavior while tracking the target velocity and path and avoiding jackknifing. Changes in road friction alter available tire forces, while terrain, load transfer, and suspension motion introduce effects that two-dimensional models struggle to capture. Prediction is challenging when these vehicle states and environmental parameters are only partially observed.

Recent approaches to handling uncertainty have focused on designing MPC schemes with learned dynamics, reducing the reliance on a fixed analytic dynamics prior \cite{learningmpc,williams2018information,kim2023bridgingactiveexplorationuncertaintyaware,NNMPC,scampicchio2025gaussianprocessesdynamicslearning}.
Another approach is policy adaptation: Rapid Motor Adaptation (RMA) augments a base policy with a latent representation of the operating conditions \cite{Kumar2021RMARM}. The policy is trained in simulation with additional environmental information not available in operating conditions. At deployment, an adaptation module infers the latent representation from recent state-action history, while the base policy weights remain fixed. This approach has also been extended to bipedal robots \cite{Kumar2022AdaptingRM}. Safe Deep Policy Adaptation combines an adaptive reinforcement learning policy with learned dynamics and a control barrier function (CBF) filter \cite{Xiao2023SafeDP}. 
We focus on a different approach: we learn the dynamics of the system directly rather than adapting a control policy online.

Learning-based MPC addresses model mismatch by learning vehicle dynamics directly from data~\cite{spielberg2019neural,NNMPC} or learning a residual correction to an analytic model~\cite{jain2021bayesrace}. Gaussian processes (GP) are commonly used for the learning framework, as they provide native uncertainty quantification output and have been shown to be effective estimators in low-data regimes \cite{scampicchio2025gaussianprocessesdynamicslearning}. More recently, however, neural dynamics have arisen as an alternative \cite{learningmpc, williams2018information, terrainaware} because they
scale to large offline datasets and can use finite histories of state--action pairs to infer latent operating conditions and mitigate partial
observability \cite{spielberg2019neural,NNMPC}. 

Recent tractor-trailer methods address forward and reverse parking using barrier-rate-guided MPPI \cite{majd2025gpu}, navigation using learned geometric distance estimates within MPPI \cite{wei2026neural}, and safety-critical trajectory planning for general \(N\)-trailer systems \cite{gao2025efficient}. High-speed studies have used analytic models for lateral stabilization and forward trajectory tracking, including active trailer steering \cite{Hou2022HighspeedLS} and tube MPC for robustness to model uncertainty and disturbances \cite{HUANG2026101493}.
Learned vehicle dynamics have also been used to improve robustness to state-estimation error and environmental variation, including in towing tasks \cite{XiaoXTKDS25}. 
Here, we address high-speed reverse tracking of a tractor-trailer system, which requires jointly tracking the target velocity and path and regulating the hitch angle under uncertain vehicle and environmental dynamics.

We propose a predictive control framework for high-speed forward and reverse tractor-trailer tracking under uncertain dynamics, combining a compact history-conditioned dynamics model with MPPI to balance velocity tracking, path following, and hitch-angle regulation. Our main contributions are:

\begin{itemize}[leftmargin=*,itemsep=1pt,topsep=2pt]

\item We integrate a learned dynamics model that takes a fixed-length state-action history as input with MPPI for high-speed forward and reverse tractor-trailer control under uncertain vehicle and environmental dynamics.

\item We derive a reduced-order tractor-trailer dynamics model using Fiala tire forces without constant longitudinal velocity assumptions to accurately characterize the forwards and reverse driving of tractor-trailers.
\item 
We demonstrate high-speed forward and reverse tracking in high-fidelity simulation across friction and terrain variations, the latter of which is absent from the training data. 

\end{itemize}

\section{PRELIMINARIES \label{sec:preliminaries}}
\subsection{Notation \label{subsec:notation}}
The sets of natural and real numbers are denoted by $\mathbb{N}$ and $\mathbb{R}$, respectively. We denote the normal distribution with mean $\mu$ and covariance $\Sigma$ by $\mathcal{N}(\mu, \Sigma)$. For a sequence \(\{\boldsymbol{a}_{n}\}\), we write \(\boldsymbol{a}_{p:q}=(\boldsymbol{a}_p,\ldots,\boldsymbol{a}_q)\), with both endpoints included. 
We use \(t\) for the current discrete control-step index, \(\tau\) for a discrete index within the prediction horizon, and \(\Delta t>0\) for the sampling interval. Furthermore, $I_{2}$ denotes the $2\times2$ identity matrix.

\subsection{System Dynamics \label{subsec:problem}}
Consider a discrete-time nonlinear system
\begin{equation}
  \boldsymbol{x}_{t+1} = f(\boldsymbol{x}_t, \boldsymbol{u}_t; \boldsymbol\vartheta)\label{eq:system}
\end{equation}
where $\boldsymbol  x_t = (\boldsymbol s_t,\boldsymbol  z_t)\in \mathcal{X}\subseteq \mathbb{R}^n$ is the state at time \(t\in \mathbb{N}\), partitioned into a measured $ \boldsymbol  s_t \in \mathbb{R}^{n_s}$ and non-measured
$\boldsymbol  z_t$. Then, \(\boldsymbol{u}_{t}\in \mathcal{U}\subset \mathbb{R}^m\) is the control input at time \(t\), and $\boldsymbol \vartheta\in \mathbb{R}^p$ collects the vehicle and environment
parameters such as mass distribution, suspension and tire characteristics,
powertrain behavior, and the friction and terrain of the track. Note that \(\vartheta\) does not vary with time.


\subsection{Model Predictive Path Integral Control\label{subsec:mppi}}

Model predictive path integral (MPPI) control is a sampling-based method for controlling nonlinear systems \cite{williams2018information,smppi}. 
At control step \(t\), let \(T\geq1\) be the prediction-horizon length, \(K\geq1\) the number of sampled control sequences, and \(\boldsymbol{u}_{t:t+T-1}\) the nominal control sequence.
MPPI samples Gaussian perturbations with positive-definite covariance \(\Sigma\in\mathbb{R}^{m\times m}\) independently across rollout steps and samples. 
The rollout commands are \(\boldsymbol{v}_{\tau}^{(k)}=\operatorname{clip}_{\mathcal{U}}(\boldsymbol{u}_{\tau}+\boldsymbol{\epsilon}_{\tau}^{(k)})\), where \(\boldsymbol{\epsilon}_{\tau}^{(k)}\sim\mathcal{N}(\boldsymbol{0},\Sigma)\), \(\tau=t,\ldots,t+T-1\), and \(k=1,\ldots,K\). Here $\boldsymbol{u}_{\tau}$ is the nominal control sequence at prediction step $\tau$ and \(\operatorname{clip}_{\mathcal{U}}\) applies the componentwise input bounds. 
Each sampled rollout sequence is computed with the next state predictor \(\tilde f\), initialized at \(\tilde{\boldsymbol{s}}_t^{(k)}=\boldsymbol{s}_t\). The \(k\)-th trajectory satisfies
\begin{equation}
    \tilde{\boldsymbol{s}}_{\tau+1}^{(k)}
    = \tilde{f}(\tilde{\boldsymbol{s}}_{\tau}^{(k)},\boldsymbol{v}_{\tau}^{(k)}),
    \quad \tau=t,\ldots,t+T-1.
\end{equation}

Let \(C_j:\mathcal{S}\times\mathcal{U}\to\mathbb{R}\) be the stage cost at time index \(j=\tau-t\) within the prediction horizon. The rollout cost is
\begin{equation*}
S^{(k)}=\sum_{\tau=t}^{t+T-1}\Bigl[
C_{\tau-t}(\tilde{\boldsymbol s}_{\tau}^{(k)},\boldsymbol v_{\tau}^{(k)}) 
+\gamma\,\boldsymbol u_{\tau}^{\top}\Sigma^{-1}
(\boldsymbol v_{\tau}^{(k)}-\boldsymbol u_{\tau})\Bigr].
\end{equation*}
where the second term is the importance-sampling correction, scaled by the tunable parameter \(\gamma\). The normalized sampling weights are
\begin{align}
    \eta &= \sum_{k=1}^K \exp\left( -\frac{1}{\lambda}\left( S^{(k)} - \min_{1\leq\ell\leq K} S^{(\ell)} \right) \right) \\
    w^{(k)} &= \frac{1}{\eta} \exp\left( -\frac{1}{\lambda}\left( S^{(k)} - \min_{1\leq\ell\leq K} S^{(\ell)} \right) \right)
    \label{eq:mppi_weights}
\end{align}
where \(\lambda>0\) is the sampling temperature. 
The control sequence is updated as \(\boldsymbol u_\tau\leftarrow\sum_{k=1}^K w^{(k)}\boldsymbol v^{(k)}_\tau\). Its first input is applied to the plant, and the remaining sequence is shifted to initialize the next control cycle.


\subsection{Problem Formulation}\label{subsec:formulation}


At each control step, we aim to solve the following receding-horizon
control problem, for $\tau=t,\ldots,t+T-1$:
\begin{equation}
\begin{aligned}
\vu_{t:t+T-1}^{\star} =
&\argmin_{\boldsymbol u_{t:t+T-1}\in\mathcal U^T}
\sum_{\tau=t}^{t+T-1}C_{\tau-t}(\tilde{\boldsymbol s}_\tau,\boldsymbol u_\tau)\\
\text{s.t.}\quad
\tilde{\boldsymbol s}_{t}=\boldsymbol s_t,\ 
&\widetilde{\mathcal H}_{t}=\mathcal H_t,
\ \tilde{\boldsymbol s}_{\tau+1}
=\tilde f_{\boldsymbol\theta}(\widetilde{\mathcal H}_{\tau},\boldsymbol u_\tau).
\end{aligned}
\label{eq:control_problem}
\end{equation}
where $\widetilde{\mathcal{H}}_{t}$ is the state-action history
of length $H$ defined as
\begin{equation}
\begin{aligned}
\widetilde{\mathcal H}_t&:=\bigl(\boldsymbol s_{t-H+1:t},\boldsymbol u_{t-H+1:t-1}\bigr)
\end{aligned}
\label{eq:history_predictor}
\end{equation}
and \(\tilde{f}_{\boldsymbol{\theta}}\) is the internal dynamics representation of the controller. We solve~\eqref{eq:control_problem} using MPPI with fixed parameters $\boldsymbol{\theta}$. In particular, we learn $\tilde{ f}_{\boldsymbol\theta}$ from an offline dataset $\mathcal{D}$ of observed state-action trajectories from~\eqref{eq:system} to support high-speed forward and reverse driving when \(f\), \(\boldsymbol \vartheta\), and \(\boldsymbol z_t\) are unavailable. We evaluate whether the learned predictor improves prediction accuracy and closed-loop performance relative to the analytic model.

\begin{figure}[t]
\centering
\includegraphics[width=0.90\linewidth]{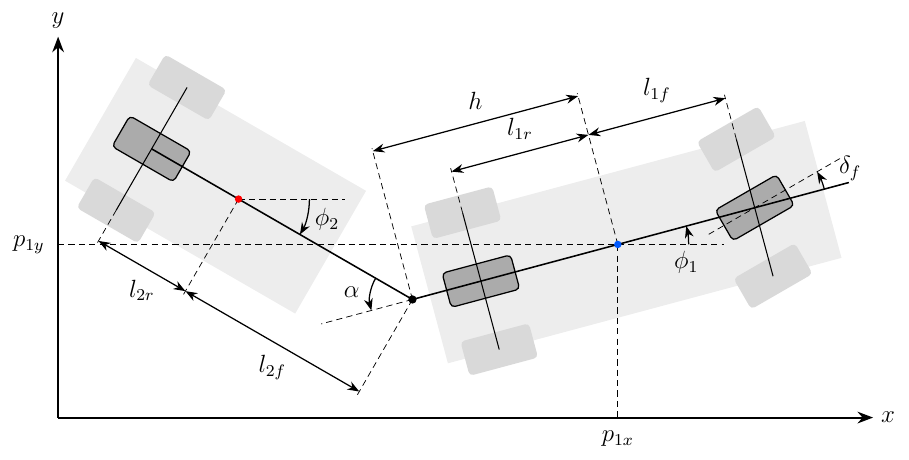}
\caption{
Illustration of our tractor-trailer model. The tractor is approximated by a bicycle model and the trailer by a single axle connected to the tractor at a common hitch point.}
\label{fig:trailer}
\vspace{-5pt}
\end{figure}



\section{METHODOLOGY }\label{sec:method}

\subsection{Tractor-trailer Dynamics\label{subsec:dyn}}

We first derive an analytic baseline of planar tractor-trailer dynamics with Fiala tire forces to accurately characterize the coupled longitudinal, lateral, yaw, and hitch motion in both forward and reverse driving while remaining low-dimensional enough for sampling-based MPC.
Our model is similar to \cite{tt_model}, but we do not have the simplifying assumption of constant longitudinal velocity, which does not hold in practical driving scenarios~\cite{design_nlmpc}. 

We approximate the tractor by a bicycle model and the trailer by a single axle, as shown in Fig.~\ref{fig:trailer}. We differentiate quantities between the tractor and trailer using a subscript $1$ or $2$ respectively. For body \(i\), let \(\phi_i\) be its yaw angle, \(m_i\) be its mass, and \(I_{zi}\) be its yaw moment of inertia about its center of mass. Let  \(l_{1f}\) and \(l_{1r}\) denote the distance from the tractor's center of mass to its front and rear axles respectively, while \(l_{2f}\) and \(l_{2r}\) run from the trailer's center of mass to the hitch and its modeled axle. Let $h$ be the distance from the tractor's center of mass to the hitch, \(\alpha:=\phi_1-\phi_2\) be the hitch angle, and
\(\delta_f\) be the physical front-wheel steering angle. 

The tractor's center-of-mass position is $\boldsymbol{p}_{1}:=(p_{1x},p_{1y})$, and \(v_{1x},v_{1y}\) are its longitudinal and lateral body-frame velocities. We define \(\boldsymbol{f}_{1f}:=(f_{1xf},f_{1yf})\) as the tractor's front tire forces in the front-wheel frame, \(\boldsymbol{f}_{1r}:=(f_{1xr},f_{1yr})\) as the tractor's rear tire forces in the body frame, and \(\boldsymbol{f}_{2r}:=(f_{2xr},f_{2yr})\) as the trailer's tire forces in the trailer frame.

Let $\boldsymbol{f}_h$ be the force the trailer exerts on the tractor at the hitch, expressed in the tractor frame, $\boldsymbol{f}_1:=\boldsymbol{f}_{1r}+R_{\delta_f}\boldsymbol{f}_{1f}=(f_{1x},f_{1y})$ the total tractor tire force, $\tau_1:=-l_{1r}f_{1yr}+l_{1f}(R_{\delta_f}\boldsymbol{f}_{1f})_y$ its yaw moment, and $\boldsymbol{w}:=\boldsymbol{v}_1-h\dot\phi_1\boldsymbol{e}_y$ the tractor's velocity at the hitch, with $\boldsymbol{e}_y:=(0,1)$ and $J:=\left[\begin{smallmatrix}0&-1\\1&0\end{smallmatrix}\right]$. The Newton--Euler equations of both bodies and the hitch constraint are
\begin{align}
\begin{split}
m_1(\dot{\boldsymbol{v}}_1+\dot\phi_1J\boldsymbol{v}_1)&=\boldsymbol{f}_1+\boldsymbol{f}_h,\\
I_{z1}\ddot\phi_1&=\tau_1-hf_{hy},
\end{split}\label{eq:ne1}\\
\begin{split}
m_2(\dot{\boldsymbol{v}}_2+\dot\phi_2J\boldsymbol{v}_2)&=\boldsymbol{f}_{2r}-R_\alpha\boldsymbol{f}_h,\\
I_{z2}\ddot\phi_2&=-l_{2r}f_{2yr}-l_{2f}(R_\alpha\boldsymbol{f}_h)_y,
\end{split}\label{eq:ne2}\\
\boldsymbol{v}_2&=R_\alpha\boldsymbol{w}-l_{2f}\dot\phi_2\boldsymbol{e}_y.\label{eq:hitch}
\end{align}
The system has an algebraic loop: $\dot{\boldsymbol{v}}_1$ depends on $\boldsymbol{f}_h$, which by \eqref{eq:ne2} depends on $\dot{\boldsymbol{v}}_2$, which depends on $\dot{\boldsymbol{v}}_1$ through \eqref{eq:hitch}. Differentiating \eqref{eq:hitch} with $\frac{d}{dt}R_\alpha=\dot\alpha JR_\alpha$ closes this loop,
\begin{equation}
\dot{\boldsymbol{v}}_2=R_\alpha\dot{\boldsymbol{w}}+\dot\alpha JR_\alpha\boldsymbol{w}-l_{2f}\ddot\phi_2\boldsymbol{e}_y,
\label{eq:hitchdot}
\end{equation}
which is affine in $\dot{\bm\nu}$. We solve the translational component of \eqref{eq:ne2} for $R_\alpha\boldsymbol{f}_h$, substitute \eqref{eq:hitchdot}, and insert the result into \eqref{eq:ne1} and the rotational component of \eqref{eq:ne2}. This eliminates $\boldsymbol{f}_h$ and leaves a system that is linear in the accelerations, $M(\alpha)\dot{\bm\nu}=\boldsymbol{b}$, with
{\small
\begin{gather}
M(\alpha)=\begin{bmatrix}
\bar m & 0 & 0 & -\rho s\\
0 & \bar m & -m_2h & -\rho c\\
0 & -m_2h & I_{z1}+m_2h^2 & \rho hc\\
-\rho s & -\rho c & \rho hc & I_{z2}+m_2 l_{2f}^2
\end{bmatrix},\nonumber\\
\boldsymbol{b}=\begin{bmatrix}
f_{1x}+\hat f_{2x}+\dot\phi_1(m_1v_{1y}+m_2\tilde v_{2y})+gc\\
f_{1y}+\hat f_{2y}-\dot\phi_1(m_1v_{1x}+m_2\tilde v_{2x})-gs\\
\tau_1-h(\hat f_{2y}-m_2\dot\phi_1\tilde v_{2x}-gs)\\
\rho\dot\phi_1v_{2x}-(l_{2f}+l_{2r})f_{2yr}
\end{bmatrix},\label{eq:dyn}
\end{gather}
}%
where $\hat{\boldsymbol{f}}_2:=R_\alpha^{\!\top}\boldsymbol{f}_{2r}$ and $\tilde{\boldsymbol{v}}_2:=R_\alpha^{\!\top}\boldsymbol{v}_2$ are the trailer force and velocity in the tractor frame and $g:=\rho\dot\alpha\dot\phi_2$. Since $\tfrac12\bm\nu^{\top}M(\alpha)\bm\nu$ is the kinetic energy, $M(\alpha)$ is symmetric positive definite for every $\alpha$, so \eqref{eq:dyn} has a unique solution.

Lateral tire forces are modeled using the Fiala brush model,
{\small
\begin{equation}
f_{y}=\begin{cases}
-C_c\tan\beta+\dfrac{C_c^{2}\lvert\tan\beta\rvert\tan\beta}{3\mu f_{z}}
-\dfrac{C_c^{3}\tan^{3}\beta}{27\mu^{2}f_{z}^{2}}, & \lvert\beta\rvert<\beta_{\mathrm{sl}},\\[4pt]
-\mu f_{z}\operatorname{sgn}\beta, & \text{otherwise},
\end{cases}
\label{eq:fiala}
\end{equation}
}
where \(C_c\in\{C_{1f},C_{1r},C_{2r}\}\) is the cornering stiffness of the corresponding axle, \(f_z\) is its static normal load, including the trailer's tongue load, and \(\mu\) is the tire--road friction coefficient, which the analytic model fixes to \(\hat\mu\). The saturation slip angle is \(\beta_{\mathrm{sl}}=\arctan(3\mu f_z/C_c)\), and the wheel slip angle is \(\beta=\operatorname{atan2}(v_\perp,\max(\lvert v_\parallel\rvert,\epsilon))\), where \(v_\parallel\) and \(v_\perp\) are wheel-frame velocities and \(\epsilon>0\) regularizes the low-speed denominator.

The tractor is rear-wheel driven. Given the normalized commands $u_\delta, u_a \in [-1,1]$, the steering angle is $\delta_f=\delta_{\max}u_\delta$, and the rear longitudinal force tracks the commanded acceleration $a_{\mathrm{cmd}}=a_{\max}\max(u_a,0)-b_{\max}\max(-u_a,0)$ up to the friction limit,
\begin{equation}
f_{1xr}=\mu f_{z}\tanh\!\big(m_1 a_{\mathrm{cmd}}/(\mu f_{z})\big),
\end{equation}
with $f_{1xf}=f_{2xr}=0$. At the rear axle, the saturation force $\mu f_z$ in \eqref{eq:fiala} is replaced by $\sqrt{(\mu f_z)^2-f_{1xr}^2}$, so throttle and braking reduce the available lateral force. Combining \eqref{eq:dyn} with the planar kinematics gives
\begin{equation}
\dot{\boldsymbol{q}} =\begin{bmatrix}
R_{\phi_1}\boldsymbol{v}_1&
\dot{\phi}_1&
\dot{\phi}_2&
M(\alpha)^{-1}\boldsymbol{b}
\end{bmatrix}^\top,
\end{equation}
where \(\boldsymbol{q} :=(p_{1x},p_{1y},\phi_1,\phi_2,\bm{\nu}^{\top})^{\top}\) is the state of the analytic model. 

\begin{figure*}[t]
\centering
\includegraphics[width=0.92\linewidth]{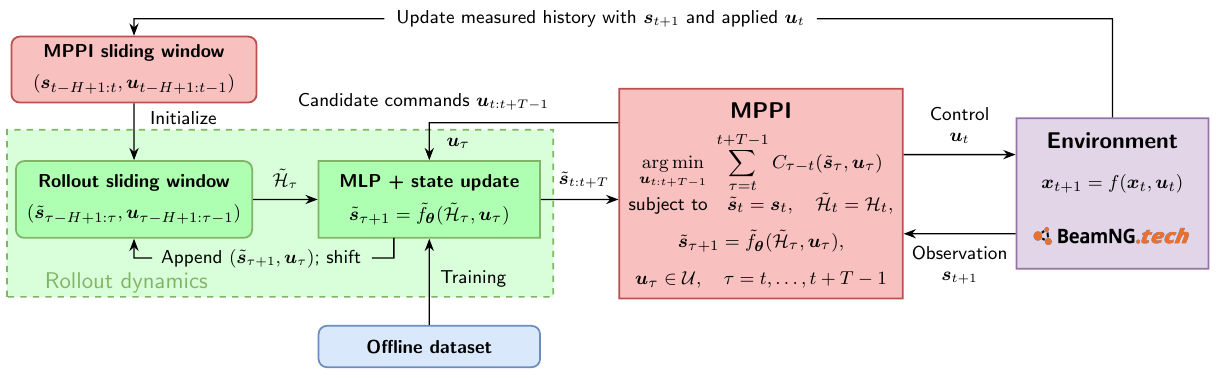}
\caption{
Overview of our controller architecture. MPPI uses a multilayer perceptron (MLP) trained offline to predict dynamic-state and actuator rates from a sliding window of state-action pairs. 
These predictions are integrated recursively to generate candidate trajectories for receding-horizon control.}
\label{fig:block}
\vspace{-5pt}
\end{figure*}

\subsection{Controller Architecture\label{subsec:arch}}
We integrate a learned dynamics model with MPPI for high-speed forward and reverse tractor-trailer control under uncertain vehicle and environmental dynamics. The model jointly predicts longitudinal, lateral, yaw, and actuator responses from recent observations and commands. It is trained offline, while its predictions maps the most recent $H$ steps of $\mathcal I_t$, $\mathcal H_t:=(\boldsymbol s_{t-H+1:t},\boldsymbol u_{t-H+1:t-1})$, and a candidate command $\boldsymbol u_t$ to $\tilde{\boldsymbol s}_{t+1}=\tilde f_{\boldsymbol\theta}(\mathcal H_t,\boldsymbol u_t)$.

\subsubsection{Dataset Collection}
We collect trajectories in the BeamNG.tech simulator on a flat asphalt environment with no snow, slope, or
banking variation. We collect uniform friction $\mu \in \{0.2, 0.4, \dots, 1.0\}$, recording
simulator states at time intervals $\Delta t = 0.05$\,s without added measurement noise.
Forward data are collected from analytic-model MPPI at
$v^\star \in \{20, 30, \dots, 90\}$\,km/h tracking a sinusoidal trajectory along the environment to better explore vehicle handling limits with added random control input noise. More specifically, we track lateral centerline offset $e_{\mathrm{ref}}(\varsigma) = \tfrac{3}{8} W \sin(\varsigma / W)$
along arc length $\varsigma$, with applied input
$u^{\mathrm{app}}_t = \mathrm{clip}_{\mathcal{U}}(u^{\mathrm{MPPI}}_t + w_t)$,
$w_t \sim \mathcal{N}(0, 0.09 I_2)$, where \(u^{\mathrm{MPPI}}_{t}\) denotes the MPPI control output at time \(t\). Reverse data are collected similarly, tracking $v^\star \in \{-20, \dots, -50\}$\,km/h but without sinusoidal tracking or input noise. We also augment the dataset with left--right mirroring. To cover large hitch angles, we add reverse runs at
$v^\star \in \{ -60,-50,-40 \}$\,km/h from MPPI using intermediate learned models.

\subsubsection{Model Training}
Our standard model is a multilayer perceptron (MLP) that predicts observed-state dynamics from \(H=4\) state-action pairs. We explore other history lengths in Section~\ref{subsec:history_comparison}. For brevity, we write \(v_x=v_{1x}\) and \(v_y=v_{1y}\). The model takes input
\[
(\sin\alpha,\cos\alpha,v_x,v_y,\dot\phi_1,\dot\phi_2,
\delta,a,u_\delta,u_a)_{t-H+1:t}
\]
and predicts the finite-difference rates
\[
(\dot v_x,\dot v_y,\ddot\phi_1,\ddot\phi_2,\dot\delta,\dot a)_t.
\]

Here \(\boldsymbol u=(u_\delta,u_a)\in\mathcal U=[-1,1]^2\) contains normalized steering and longitudinal actuation commands, with observed responses \(\delta,a\). 
At the start of each rollout, we initialize the history with the $H$ most recent observations and the $H-1$ most recently applied commands. We append the first command from the sampled sequence to the command history to form the predictor input. The magnitude of $u_{a}$ denotes the throttle command if $u_{a}$ is positive and denotes braking when negative. The actuator response $a$ is computed as the realized throttle minus the realized braking. Denote the normalized steering angle by $\delta$ and the physical steering angle by $\delta_{f}$. We use finite differences of the measured velocities, yaw rates, and actuator states for training.

During rollouts, we integrate the predicted rates with forward Euler to update these states and use the vehicle kinematics to update position and orientation.
The MLP has four hidden layers of 128 neurons each. Inputs and targets are standardized using train set means and standard deviations.

\subsubsection{MPPI Controller and Cost Function}
Each MPPI rollout initializes its own history from measurements, then updates its window with predicted states and sampled controls (Fig.~\ref{fig:block}). Every rollout computes a cost that promotes target-velocity tracking and penalizes track-boundary and hitch-angle violations. The violation penalty \(c_{\text{viol}}\) is
\begin{align}
 c_{\text{viol}}&=c_{\text{track}}+c_{\text{jackknife}} \label{cviol} \\
 c_{\text{track}}&=\max\left(0, \lvert e_{\mathrm{lat}} \rvert - 0.45 W_\mathrm{track} + 0.1\right) \nonumber\\
 c_{\text{jackknife}}&=\max(0, \lvert\alpha\rvert - \alpha_{\mathrm{max}}) \nonumber
\end{align}
where $e_{\mathrm{lat}}$ is the signed lateral deviation of the tractor's center of mass from the track centerline, $\alpha_{\mathrm{max}}$ is the prescribed jackknife threshold, and $W_{\mathrm{track}}$ is the track width. Distances in~\eqref{cviol} are measured in meters, so the constant $0.1$ represents a $0.1\,\mathrm{m}$ margin. When $\lvert\alpha\rvert>\alpha_{\mathrm{max}}$, we consider the tractor-trailer to have jackknifed. Additionally, we define 
\begin{align}
    c_{\text{vel}}&=\lvert v^\star-v_{\mathrm{track}}\rvert \label{cvel} \\
    c_{\text{center}} & = e_{\text{lat}}^2 \label{ccenter} \\
    c_{\text{hitch}} & = \alpha^2  \label{chitch}
\end{align}
where $v^\star$ is the signed target velocity and $v_{\mathrm{track}}$ is the signed velocity along the track centerline. Combining \eqref{cviol}--\eqref{chitch}, 
the stage cost at time index $j=\tau-t\in\{0,\ldots,T-1\}$ within the prediction horizon is
\begin{equation}
\begin{aligned}
C_j(\boldsymbol s,\boldsymbol u)
={}&0.99^j\cdot10^7c_{\text{viol}}+k_vc_{\text{vel}}\\
&+k_cc_{\text{center}}+k_ac_{\text{hitch}}.
\end{aligned}
\label{eq:costform}
\end{equation}
where $k_{v}$, $k_{c}$, and $k_{a}$ are penalty weights.

The \(K\) rollout costs update the control sequence to approximately solve~\eqref{eq:control_problem}, and the first command is applied. This command and the new observation update the measured history for replanning. Main tracking experiments initialize the history with \(H\) steps of analytic-model MPPI.

\begin{figure*}[t]
\centering
\includegraphics[width=0.9\linewidth]{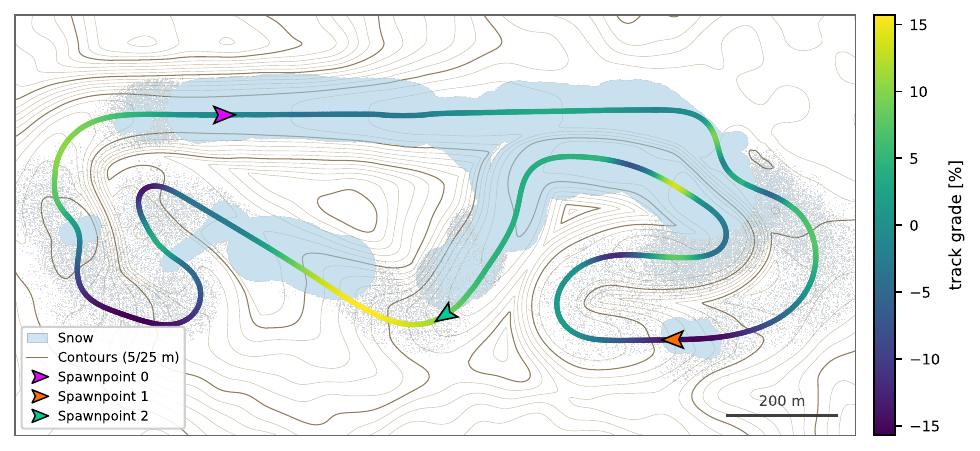}
\caption{Top-down view of the Snow + Slope test environment, based on the Circuit de Barcelona-Catalunya layout. Snow patches and terrain changes are applied to the base asphalt track. The snow pattern matches the Snow environment, and the terrain profile matches the Slope environment.}
\label{fig:env}
\vspace{-5pt}
\end{figure*}

\section{SIMULATION RESULTS}\label{sec:results}

We address three research questions:
\begin{enumerate}[label=\textbf{Q\arabic*:},leftmargin=*,itemsep=1pt,topsep=2pt]
    \item Does MPPI with learned dynamics improve forward and reverse tracking relative to MPPI with the analytic model across friction and terrain variations?     
    \item How do learned and analytic dynamics compare in trajectory rollout prediction, and does exposing the true friction coefficients close the gap?
    \item How does closed-loop performance compare among selected learned predictors with history lengths $H\in\{1,4,8\}$?
\end{enumerate}

\subsection{Experimental Setup}\label{subsec:experimental_setup}
\label{subsec:env}\label{subsec:baselines}
\textbf{Simulation environments:} We evaluate the controllers in BeamNG.tech \cite{beamng_tech}, whose internal plant dynamics are unavailable to the controllers. 
The inputs are $\boldsymbol u=(u_\delta,u_a)\in[-1,1]^2$, and the observed state is $\boldsymbol s=(p_{1x},p_{1y},\phi_1,\phi_2,v_{1x},v_{1y},\dot\phi_1,\dot\phi_2,\delta,a)\in\mathbb R^{10}$. The unobserved $z_t$ contains state information including vehicle pitch and roll, suspension states, and powertrain states, while vehicle parameters, friction, and terrain enter through $\vartheta$. The tractor-trailer follows a predefined racetrack at high target speeds in forward and reverse motion, providing a stress test of dynamics prediction. The track follows the general layout of the Circuit de Barcelona-Catalunya. Unobserved plant states \(\boldsymbol{z}_t\) include vehicle pitch and roll, suspension states, local road friction and roughness, and powertrain behavior.

The four test environments are:
\begin{enumerate}[leftmargin=*,itemsep=1pt,topsep=2pt,parsep=0pt]
    \item[1.] \textit{Asphalt}: A flat track with uniformly high-friction asphalt.
    \item[2.] \textit{Snow}: A flat asphalt track with irregularly distributed snow patches that vary the local friction.
    \item[3.] \textit{Slope}: A uniformly high-friction track with elevation changes and banking.
    \item[4.] \textit{Snow + Slope}: The same asphalt-snow pattern as in
    the Snow scenario, with the same terrain pattern as the Slope scenario.
\end{enumerate}
Figure~\ref{fig:env} shows the combined snow and terrain modifications.

\textbf{Controllers:} We compare the following five controllers:
\begin{enumerate}[leftmargin=*,itemsep=1pt,topsep=2pt,parsep=0pt]
    \item \textit{Controller 1 (Ours)}: An MPPI controller using the learned dynamics model described in Section~\ref{subsec:arch}.
    \item \textit{Controller 2}: An MPPI controller with the analytic model described in Section~\ref{subsec:dyn}.
    \item \textit{Controller 3}: A hybrid controller that uses a linear quadratic regulator (LQR) \cite{lqr} for steering and proportional--integral--derivative (PID) control \cite{PID} for longitudinal actuation. The LQR controller operates on a linearization of the analytic dynamics model at the \(\alpha=0\) operating point and assumes \(v_y=0\) and constant \(v_x\).
    \item \textit{Controller 4}: A gradient-based MPC controller that uses IPOPT \cite{IPOPT} with direct shooting under the analytic dynamics model.

    \item \textit{Controller 5}: A gradient-based model predictive contouring control (MPCC) controller \cite{mpcc} using our analytic dynamics model.
\end{enumerate}

\begin{figure*}[t]
\centering
\includegraphics[width=0.99\linewidth]{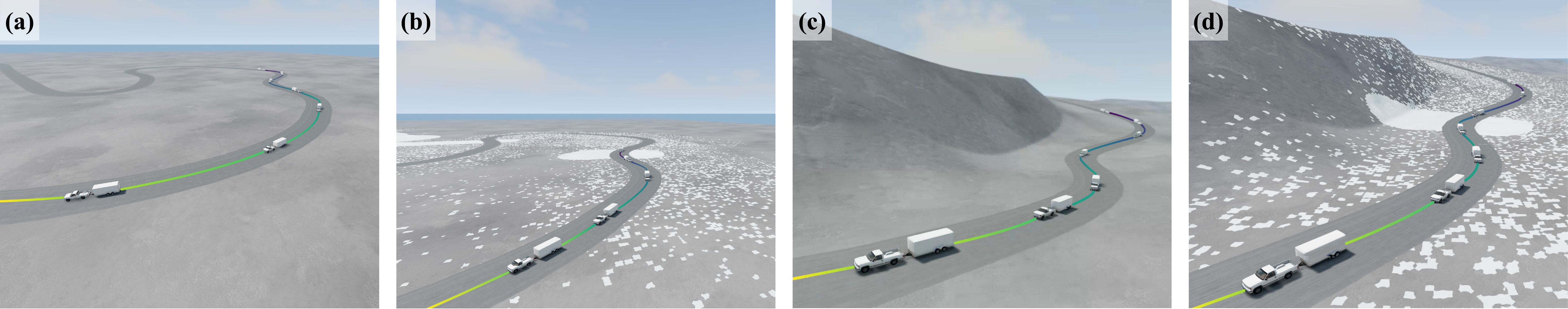}
\caption{Representative reverse-driving trajectories of Controller 1 in (a) Asphalt, (b) Snow, (c) Slope, and (d) Snow + Slope. White patches indicate snow-covered regions with lower friction. }
\label{fig:scenarios}
\vspace{-5pt}
\end{figure*}

\subsection{Closed-Loop Tracking \label{subsec:setup}}

Figure~\ref{fig:scenarios} shows representative reverse-driving trajectories of Controller 1 in the four environments defined in Section~\ref{subsec:experimental_setup}.
Table~\ref{tab:mppi_tuning} gives the MPPI settings for these main tracking experiments.
\begin{table}[htpb]
\centering
\caption{MPPI settings for the main tracking experiments. Sampling covariance is expressed in normalized command coordinates $(u_\delta,u_a)$.}
\label{tab:mppi_tuning}
\small
\begin{tabular}{lcc}
\hline
\textbf{Parameter} & \textbf{Forward} & \textbf{Reverse} \\
\hline
$k_{v}$        & $1\times10^{2}$ & $5\times10^{1}$ \\
$k_{c}$        & $1\times10^{0}$ & $5\times10^{1}$ \\
$k_{a}$        & $7\times10^{2}$ & $1\times10^{2}$ \\
$\lambda$ (temperature)     & $150$           & $100$           \\
$\operatorname{diag}(\Sigma)$    & $(7\times10^{-2},\,0.2)$ & $(2\times10^{-2},\,0.2)$ \\
$K$ (samples)              & $500$           & $500$          \\
$T$ (horizon steps)              & $80$            & $65$            \\
$\gamma$                   & $0.0$           & $0.0$           \\
$\Delta t$ (s)          & $0.05$          & $0.05$          \\
\hline
\end{tabular}
\end{table}
Table~\ref{tab:sweep} summarizes performance in the four scenarios.
\begin{table*}[t]
\centering

\caption{Closed-loop tracking across four environments, with nine runs per configuration. 
Controllers are defined in Section~\ref{subsec:experimental_setup}. Bold hitch-angle values are the lowest for each velocity setpoint and environment.}
\label{tab:sweep}
\setlength{\tabcolsep}{3pt}
\renewcommand{\arraystretch}{1.05}
\resizebox{\linewidth}{!}{%
\begin{tabular}{c r rrrc rrrc rrrc rrrc}
\toprule
& & \multicolumn{4}{c}{Asphalt} & \multicolumn{4}{c}{Snow}
    & \multicolumn{4}{c}{Slope} & \multicolumn{4}{c}{Snow + Slope} \\
\cmidrule(lr){3-6}\cmidrule(lr){7-10}\cmidrule(lr){11-14}\cmidrule(lr){15-18}
Controller & $v^\star$
 & $\bar v$ & $\bar\alpha_{\mathrm{rms}}$ & $\bar N$ & C/J/S/O/F
 & $\bar v$ & $\bar\alpha_{\mathrm{rms}}$ & $\bar N$ & C/J/S/O/F
 & $\bar v$ & $\bar\alpha_{\mathrm{rms}}$ & $\bar N$ & C/J/S/O/F
 & $\bar v$ & $\bar\alpha_{\mathrm{rms}}$ & $\bar N$ & C/J/S/O/F \\
 & {\footnotesize (km/h)}
 & {\footnotesize (km/h)} & {\footnotesize (deg)} & {\footnotesize (steps)} & {\footnotesize (out of 9)}
 & {\footnotesize (km/h)} & {\footnotesize (deg)} & {\footnotesize (steps)} & {\footnotesize (out of 9)}
 & {\footnotesize (km/h)} & {\footnotesize (deg)} & {\footnotesize (steps)} & {\footnotesize (out of 9)}
 & {\footnotesize (km/h)} & {\footnotesize (deg)} & {\footnotesize (steps)} & {\footnotesize (out of 9)} \\
\midrule

\multirow{3}{*}{\shortstack[l]{1: \textbf{Ours}}} & $-30$ & 29.1 & \textbf{1.14} & \textbf{500} & 9/0/0/0/0 & 29.1 & \textbf{1.18} & \textbf{500} & 9/0/0/0/0 & 25.4 & \textbf{1.03} & 455 & 8/0/1/0/0 & 20.4 & \textbf{10.82} & 386 & 6/0/3/0/0 \\
 & $-50$ & 47.2 & \textbf{1.75} & \textbf{500} & 9/0/0/0/0 & 46.0 & \textbf{1.82} & \textbf{500} & 9/0/0/0/0 & 42.1 & \textbf{3.06} & 498 & 8/1/0/0/0 & 30.4 & \textbf{15.41} & 413 & 4/0/5/0/0 \\
 & $80$ & 75.4 & 2.57 & \textbf{500} & 9/0/0/0/0 & 72.6 & 2.72 & \textbf{500} & 9/0/0/0/0 & 70.7 & 2.67 & \textbf{500} & 9/0/0/0/0 & 69.1 & 2.83 & \textbf{500} & 9/0/0/0/0 \\
\midrule

\multirow{3}{*}{\shortstack[l]{2: MPPI (analytic)} \cite{williams2018information}} & $-30$ & 19.7 & 5.25 & 392 & 5/0/4/0/0 & 19.6 & 7.82 & 403 & 5/0/4/0/0 & 22.3 & 2.86 & 498 & 8/0/1/0/0 & 16.7 & 11.47 & 365 & 4/0/5/0/0 \\
 & $-50$ & 22.7 & 6.70 & 448 & 6/0/3/0/0 & 22.8 & 8.49 & 417 & 5/0/4/0/0 & 22.5 & 6.48 & 457 & 6/0/3/0/0 & 18.1 & 15.86 & 321 & 2/1/6/0/0 \\
 & $80$ & 71.1 & 3.00 & \textbf{500} & 9/0/0/0/0 & 68.7 & 3.26 & \textbf{500} & 9/0/0/0/0 & 67.1 & 3.19 & \textbf{500} & 9/0/0/0/0 & 64.8 & 3.66 & \textbf{500} & 9/0/0/0/0 \\
\midrule
\multirow{3}{*}{3: LQR--PID \cite{lqr, PID}}  & $-30$ & 30.8 & 4.80 & 495 & 8/0/0/1/0 & 30.6 & 5.36 & 439 & 6/0/0/3/0 & 28.5 & 10.02 & 446 & 6/0/0/3/0 & 24.0 & 16.56 & 377 & 3/6/0/0/0 \\
 & $-50$ & 45.1 & 14.41 & 452 & 6/1/0/2/0 & 48.8 & 8.97 & 407 & 6/3/0/0/0 & 46.9 & 10.58 & 396 & 6/3/0/0/0 & 37.1 & 16.91 & 352 & 3/3/0/3/0 \\
 & $80$ & 74.8 & \textbf{2.45} & 436 & 6/0/0/3/0 & 72.4 & \textbf{2.36} & 442 & 6/0/0/3/0 & 71.7 & \textbf{2.52} & \textbf{500} & 9/0/0/0/0 & 70.7 & \textbf{2.53} & \textbf{500} & 9/0/0/0/0 \\
\midrule
\multirow{3}{*}{4: Shooting MPC \cite{IPOPT}} & $-30$ & 28.4 & 3.11 & \textbf{500} & 9/0/0/0/0 & 23.5 & 20.22 & 373 & 5/1/3/0/0 & 19.9 & 16.32 & 478 & 7/2/0/0/0 & 16.7 & 20.79 & 338 & 4/0/5/0/0 \\
 & $-50$ & 32.2 & 54.93 & 107 & 0/9/0/0/0 & 31.9 & 39.96 & 244 & 0/6/0/3/0 & 26.3 & 28.09 & 293 & 3/3/3/0/0 & 23.5 & 30.59 & 292 & 3/3/3/0/0 \\
 & $80$ & 71.3 & 12.28 & 471 & 5/1/0/3/0 & 69.1 & 9.16 & 468 & 6/3/0/0/0 & 69.5 & 8.13 & 490 & 6/0/0/3/0 & 66.7 & 11.80 & 475 & 6/0/0/3/0 \\
\midrule
\multirow{3}{*}{5: MPCC \cite{mpcc}} & $-30$ & 22.0 & 26.57 & 301 & 0/3/2/4/0 & 21.9 & 39.16 & 261 & 0/3/3/3/0 & 18.1 & 30.46 & 374 & 1/0/7/0/1 & 14.1 & 31.79 & 275 & 0/1/6/2/0 \\
 & $-50$ & 39.3 & 14.74 & 261 & 0/3/1/4/1 & 30.2 & 24.95 & 203 & 0/2/3/4/0 & 37.3 & 17.98 & 352 & 1/3/1/3/1 & 25.5 & 25.63 & 202 & 0/1/3/5/0 \\
 & $80$ & 68.0 & 6.06 & 400 & 4/0/0/5/0 & 66.1 & 4.82 & 407 & 4/0/0/4/1 & 64.5 & 6.73 & 379 & 3/0/0/5/1 & 64.4 & 5.74 & 426 & 4/0/0/5/0 \\
\bottomrule
\end{tabular}%
}
\par
\vspace{4pt}
\parbox{\linewidth}{\footnotesize
Each configuration comprises three seeds at each of three locations
($n=9$). Our notation is as follows: $v^\star$ is the signed velocity setpoint, with negative values denoting reverse motion. $\bar v$ and $\bar\alpha_{\mathrm{rms}}$ are the arithmetic means of the per-run mean speed magnitudes and RMS hitch angles, respectively, computed after the first 100 control steps. $\bar N$ is the mean number of completed control steps out of 500. All means include unsuccessful runs and weight each run equally. Bold $\bar N$ entries indicate that all nine runs completed 500 steps. C/J/S/O/F gives the numbers of completed, jackknifed, stalled, off-track, and solver-failed runs, respectively. Completion denotes reaching the 500-step (25-second) test horizon. A stall is defined by $\lvert v_x\rvert<0.5\,\mathrm{m/s}$ with no track progress for $2\,\mathrm{s}$.}
\end{table*}

\subsubsection{Asphalt Scenario\label{asphalt}}

Controller 1 is the only controller to complete all nine 500-step runs at each of the three setpoints, \(v^\star\in\{-30,-50,80\}\,\mathrm{km/h}\). It achieves the lowest RMS hitch angle at both reverse setpoints. In forward motion, Controller 3 achieves a lower RMS hitch angle but completes fewer runs. Controller 1 also achieves mean speeds closer to the target magnitudes than Controller 2 at every setpoint. At \(-50\,\mathrm{km/h}\), for example, Controller 1 achieves \(\bar v=47.2\,\mathrm{km/h}\), \(\bar\alpha_{\mathrm{rms}}=1.75^\circ\), and nine completions, compared with \(22.7\,\mathrm{km/h}\), \(6.70^\circ\), and six completions for Controller 2.

\subsubsection{Snow Scenario}
Controller 1 completes all nine Snow runs at each setpoint, with mean speeds closer to the target magnitudes than Controller 2. It has no jackknife or stall failures despite its higher reverse speeds, unlike Controller 2, which stalls in four of nine runs at each reverse setpoint. Controller 3 has higher RMS hitch angles in reverse but a slightly lower forward value ($2.36^\circ$ vs. $2.72^\circ$). Controller 4 has higher RMS hitch angles and fewer completed steps than Controller 1 at every setpoint. Controller 5 has the fewest completed steps at each setpoint and the fewest completions overall.

\subsubsection{Slope Scenario}

In the Slope environment, Controller 1 completes eight of nine runs at each reverse setpoint, compared with all nine on Asphalt and Snow. The two reverse terminations consist of one stall at \(-30\,\mathrm{km/h}\) and one jackknife at \(-50\,\mathrm{km/h}\). It still achieves mean speeds closer to the target magnitudes and lower RMS hitch angles than Controller 2 at every setpoint. It completes at least as many reverse runs and has lower reverse RMS hitch angles than every baseline. In forward motion, Controller 3 matches its completion rate with a slightly higher mean speed and lower RMS hitch angle.

\subsubsection{Snow + Slope Scenario}
Controller 1 has the most reverse completions and the lowest reverse RMS hitch angles on Snow + Slope, although stalls limit its completion rate. 
This combined environment also has the largest reduction in reverse completions for MPPI with learned dynamics: 10 of 18 runs, compared with 18 of 18 on Snow and 16 of 18 on Slope. It is the only controller with no jackknife failures across the three setpoints. Controller 3 achieves a higher mean speed and slightly lower RMS hitch angle in forward driving, but completes fewer reverse runs, with more jackknife or off-track failures.

\subsubsection{Summary Across Environments}

Across the eight reverse configurations in Table~\ref{tab:sweep}, Controller 1 completes 62 of 72 runs, compared with 41 of 72 for Controller 2. The corresponding stall counts are nine and 30, while each controller has one jackknife failure. Controller 1 also achieves higher mean speeds and lower RMS hitch angles in every reverse configuration. Both MPPI variants complete all 36 forward runs. 
LQR--PID has lower forward RMS hitch angles than MPPI with learned dynamics in all four environments, with fewer completions on Asphalt and Snow.

Despite the learned predictor not receiving terrain-map information, unlike the terrain-aware model in \cite{terrainaware} and terrain variation absent from its training data, we find that the learned model transfers to the tested terrain profiles, but combined snow and terrain variation still causes stalls, as Controller 1 completes six of nine reverse runs at \(-30\,\mathrm{km/h}\) and four of nine at \(-50\,\mathrm{km/h}\).

\begin{figure*}[t]
\centering
\captionsetup{skip=3pt}
\includegraphics[width=0.9\textwidth]{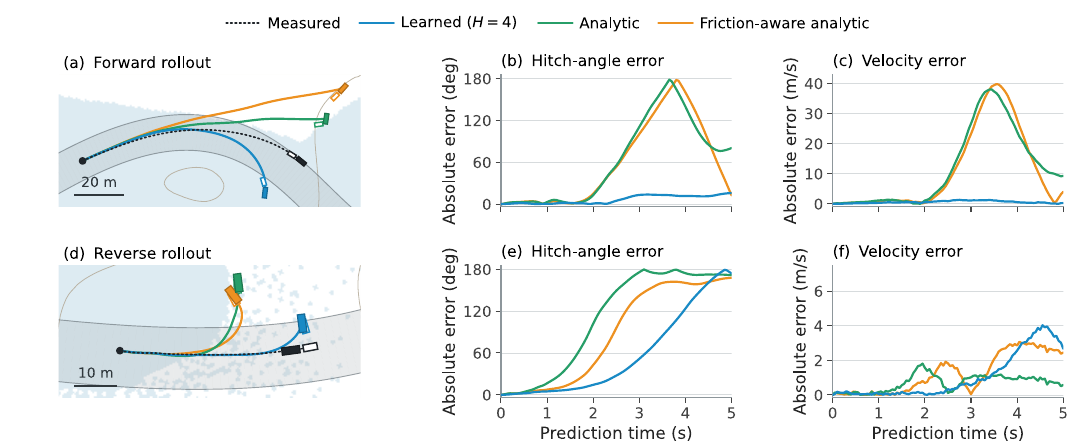}
\caption{Five-second open-loop dynamics predictions in Snow + Slope for forward (a,b,c) and reverse (d,e,f) driving. The measured trajectory is shown in dotted black, and the dynamics models' open-loop predictions are shown in solid color. Terminal tractor and trailer poses are denoted by solid and hollow boxes, respectively. The blue shading denotes snow patches with decreased friction, and elevation contours are spaced at 5\,m.}
\label{fig:divergence_snow_slope}
\vspace{-5pt}
\end{figure*}

\subsection{Dynamics Prediction Error \label{subsec:dynamics}}

We compare the multi-step dynamics prediction accuracy of the following models:
\begin{enumerate}[leftmargin=*,itemsep=1pt,topsep=2pt,parsep=0pt]
    \item[1.] The analytic model (defined in Section \ref{subsec:dyn}), with fixed friction \(\mu=1\).
    \item[2.] The analytic model, given the ground-truth tractor and trailer contact-friction coefficients.
    \item[3.] The fully learned model with \(H=4\).
\end{enumerate}

From MPPI trajectories with learned dynamics, we select one five-second BeamNG.tech segment per direction and environment. Each predictor replays the same 100 recorded commands (\(\Delta t=0.05\,\mathrm{s}\)) from the same measured state without resets, fixing actions across models; the learned predictor initializes from four measured state-action pairs. The friction-aware model receives recorded contact-friction sequences, not a map evaluated along its predicted path. Figure~\ref{fig:divergence_snow_slope} shows Snow + Slope rollouts and absolute wrapped hitch-angle and longitudinal velocity errors. Table~\ref{tab:rmse_comparison} reports position and wrapped hitch-angle RMSE over the 100 predicted states, excluding the common initial state.

\begin{table}[tbp]
\centering
\caption{Position ($e_p$) and wrapped hitch-angle ($e_\alpha$) RMSE for one five-second segment per direction and environment (meters and degrees, respectively).
Bold values indicate the lowest RMSE for each environment, direction, and metric.}
\label{tab:rmse_comparison}
\footnotesize
\setlength{\tabcolsep}{4pt}
\begin{tabular}{@{}llrrrrrr@{}}
\toprule
& & \multicolumn{2}{c}{Analytic}
  & \multicolumn{2}{c}{\shortstack{Friction-aware\\analytic}}
  & \multicolumn{2}{c}{Learned} \\
\cmidrule(lr){3-4}
\cmidrule(lr){5-6}
\cmidrule(lr){7-8}
Direction & Environment
& \(e_p\) & \(e_\alpha\)
& \(e_p\) & \(e_\alpha\)
& \(e_p\) & \(e_\alpha\) \\
\midrule
Reverse & Asphalt
& 5.8 & \textbf{96.5}
& 7.4 & 113.0
& \textbf{4.2} & 101.1 \\
& Snow
& 7.0 & 130.1
& 4.9 & 107.9
& \textbf{4.0} & \textbf{88.2} \\
& Slope
& 7.7 & 132.6
& 8.2 & 134.2
& \textbf{2.2} & \textbf{78.7} \\
& Snow + Slope
& 8.1 & 131.5
& 7.3 & 111.6
& \textbf{2.1} & \textbf{83.4} \\
\midrule
Forward & Asphalt
& 7.3 & 93.4
& 6.6 & 95.3
& \textbf{1.4} & \textbf{2.1} \\
& Snow
& \textbf{7.3} & 82.3
& 11.9 & 83.8
& 10.8 & \textbf{16.0} \\
& Slope
& 3.5 & 80.4
& \textbf{2.7} & 79.2
& 5.7 & \textbf{1.1} \\
& Snow + Slope
& 8.4 & 85.7
& 14.7 & 84.0
& \textbf{7.5} & \textbf{8.7} \\
\bottomrule
\end{tabular}
\end{table}

During reverse driving, the learned model achieves the lowest position RMSE in all four environments and the lowest hitch-angle RMSE in three of four. On Asphalt, the fixed-friction analytic has lower hitch-angle RMSE ($96.5^\circ$ versus $101.1^\circ$). Figure~\ref{fig:divergence_snow_slope}(e) shows slower initial growth of the learned model's hitch-angle error, although its error becomes large later in the five-second horizon.

During forward driving, the learned model achieves the lowest hitch-angle RMSE in all four environments. 
Across the eight selected segments, the friction-aware analytic model has lower position RMSE than the fixed-friction analytic model in four cases and lower hitch-angle RMSE in four cases. Providing the true friction value to the analytic model does not allow it to match the performance of the learned model (\textbf{Q2}) and does not consistently improve dynamics prediction. This is likely due to the effects of other unmodeled dynamics. We note that these comparisons do not establish how much of the remaining error comes from unmodeled vehicle behavior and how much comes from other environmental effects. 

The five-second diagnostic horizon also exceeds the main MPPI horizons of $4.0\,\mathrm{s}$ forward and $3.25\,\mathrm{s}$ reverse. The large late reverse errors show the limits of open-loop prediction despite the closed-loop tracking results above. Because hitch error is wrapped, a late decrease after a large trailer rotation need not indicate recovery.

\subsection{History-Length Comparison}\label{subsec:history_comparison}
We compare selected predictors with $H\in\{1,4,8\}$ on a separate Snow + Slope segment, using three seeds at one start per direction (Table~\ref{tab:history_snow_slope}). The $H=1$ model uses fixed-friction training, whereas $H=4$ and $H=8$ use mixed-friction data.

\suppressfloats[t]
\begin{table}[t]
\centering
\caption{Closed-loop history comparison on Snow + Slope.}
\label{tab:history_snow_slope}
\setlength{\tabcolsep}{3.1pt}
\renewcommand{\arraystretch}{1.09}
\footnotesize
\begin{tabular*}{\linewidth}{@{\extracolsep{\fill}}lrrrrc@{}}
\toprule
Predictor & $v^\star$ & $\bar v$ & $\bar\alpha_{\mathrm{rms}}$ & $\bar N$ & C/J/S/O/F \\
 & (km/h) & (km/h) & (deg) & (steps) & (out of 3) \\
\midrule
$H=1$ & $-30$ & 15.8 & 29.29 & 98 & 0/3/0/0/0 \\
 & $80$ & 48.0 & 22.27 & 278 & 0/2/0/1/0 \\
\midrule
$H=4$ & $-30$ & 30.0 & \textbf{2.84} & \textbf{500} & 3/0/0/0/0 \\
 & $80$ & 51.6 & 29.29 & 345 & 0/3/0/0/0 \\
\midrule
$H=8$ & $-30$ & 30.9 & 2.87 & \textbf{500} & 3/0/0/0/0 \\
 & $80$ & 64.0 & \textbf{6.07} & \textbf{500} & 3/0/0/0/0 \\
\midrule
Analytic & $-30$ & 18.9 & 13.78 & 393 & 2/0/1/0/0 \\
 & $80$ & 53.3 & 20.00 & 356 & 0/2/0/1/0 \\
\bottomrule
\end{tabular*}
\par\vspace{3pt}
\parbox{\linewidth}{\scriptsize Each model has three seeds at one start location per direction. 
$\bar v$ and $\bar\alpha_{\mathrm{rms}}$ are means of the per-run values, with equal weight assigned to each run. Each value is computed after the first 100 steps, or over the whole run if it ends within 100 steps. $\bar N$ is the mean duration out of 500 steps ($25$ s). C/J/S/O/F counts completed/jackknifed/stalled/off-track/solver-failed runs. Bold indicates minimum RMS hitch angle per direction, or all runs reaching 500 steps.}
\end{table} 

For this comparison, we use $K=1500$, $T=80/75$ for forward/reverse driving, $\lambda=100$, $\operatorname{diag}(\Sigma)=(0.01,0.2)$, $\gamma=0$, and $\Delta t=0.05\,\mathrm{s}$. The weights $(k_v,k_c,k_a)$ are $(20,50,100)$ forward and $(50,50,200)$ reverse. All candidates use these numerical MPPI settings. Both the starting segment and tuning differ from the main experiments, so completion rates should not be compared directly with Table~\ref{tab:sweep}.

The $H=1$ model completes no runs in either direction; all three reverse runs end in jackknifing, with a mean duration of 98 control steps. Both $H=4$ and $H=8$ complete all reverse runs, with similar RMS hitch angles of $2.84^\circ$ and $2.87^\circ$. The $H=4$ mean speed is also closer to the target magnitude than that of $H=8$.

Their forward outcomes differ: $H=4$ jackknifes in all three runs, whereas $H=8$ completes all three and has the lowest forward RMS hitch angle. Its mean speed of $64.0\,\mathrm{km/h}$ remains below the $80\,\mathrm{km/h}$ target. 
The forward RMS hitch angle does not decrease monotonically with $H$, as the $H=4$ value exceeds that of $H=1$. The analytic model completes two reverse runs and no forward runs. Although its forward mean speed and RMS hitch angle are better than those of $H=4$, neither completes a forward run. These results answer \textbf{Q3} for the selected predictors, but differences in their training data (fixed vs mixed friction) prevent attributing the outcomes to history length alone.

\section{CONCLUSION }\label{sec:conclusion}

We evaluated MPPI with learned tractor-trailer dynamics against a Fiala-based analytic model and additional controllers. In every tracking configuration, learned dynamics brought mean speeds closer to the target magnitudes and reduced RMS hitch angles relative to analytic-model MPPI. Reverse completions increased from 41 to 62 of 72 runs. This advantage persisted on slope and banking absent from training, although combined snow and terrain variation limited reverse completion. On a separate Snow + Slope segment, controllers using \(H=4\) and \(H=8\) predictors completed all reverse runs, but only \(H=8\) completed all forward runs. Future work should consider other sources of uncertainty, such as trailer payload variation.

\addtolength{\textheight}{0 cm}   




\bibliographystyle{IEEEtran}
\typeout{}
\bibliography{references.bib}


\end{document}